# Distributed Classification of Urban Congestion Using VANET

Ranwa Al Mallah, Alejandro Quintero, and Bilal Farooq

*Abstract*—**Vehicular ad hoc networks (VANETs) can efficiently detect traffic congestion, but detection is not enough, because congestion can be further classified as recurrent and non-recurrent congestion (NRC). In particular, NRC in an urban network is mainly caused by incidents, work zones, special events, and adverse weather. We propose a framework for the real-time distributed classification of congestion into its components on a heterogeneous urban road network using VANET. We present models built on an understanding of the spatial and temporal causality measures and trained on synthetic data extended from a real case study of Cologne. Our performance evaluation shows a predictive accuracy of 87.63% for the deterministic classification tree, 88.83% for the naïve Bayesian classifier, 89.51% for random forest, and 89.17% for the boosting technique. This framework can assist transportation agencies in reducing urban congestion by developing effective congestion mitigation strategies knowing the root causes of congestion.**

## I. INTRODUCTION

CONGESTION can be classified as recurrent and non-recurrent. Recurrent congestion refers to congestion that happens on a regular basis and usually occurs when a large number of vehicles use the limited capacity of the road simultaneously. Non-recurrent congestion (NRC) in an urban network is mainly caused by incidents, workzones, special events and adverse weather [1].

Recently, the intelligent transportation system research has shifted its focus to the next generation sensing technology, vehicular ad-hoc network (VANET). Advances in vehicle-to-vehicle (V2V) and vehicle-to-infrastructure (V2I) wireless communications have increased the potential of real-time monitoring of traffic variables in a distributed manner. Distributed monitoring refers to the process by which macroscopic and microscopic traffic variables are collected by vehicles themselves without the need of infrastructure. A large amount of works have been proposed for the congestion detection problem using VANET [2]. These approaches only detect congestion and do not clarify if it is due to recurrent or NRC.

R. Al Mallah and A. Quintero are with the Department of Computer Science, École Polytechnique de Montréal, Montréal, QC H3T 1J4, Canada, (e-mail: ranwa.al-mallah@polymtl.ca; alejandro.quintero@polymtl.ca)

B. Farooq is with the Department of Civil, Geological and Mining Engineering, École Polytechnique de Montréal, Montréal, QC H3T 1J4, Canada, (e-mail: bilal.farooq@polymtl.ca)

They cannot be used to classify the congestion into its components.

In transportation, although understanding how much of the total congestion is due to NRC has been thoroughly studied for both highway [3] and urban traffic [1], several unresolved problems still exist. Firstly, the duration, timing and location of NRC in an urban road network varies highly. Thus making it difficult to monitor traffic in real time or on a continuing basis with conventional induction loops, cameras and floating cars mechanisms which are expensive to deploy and maintain for large coverage areas. Alternative cost-effective and flexible solutions are needed to guarantee better monitoring of road traffic at various level of granularity. Secondly, existing NRC detection methods not only need extensive datasets, but they are also not deployed in real-time. In real-time, valuable information with regard to the impacts of the detected NRC can be disseminated to drivers and traffic management centers so that appropriate proactive strategies for recovering traffic conditions back to normality can be set in place. Finally, NRC detection methods should be able to better characterize a NRC event once it is detected. Existing methods only quantify the spatial and temporal impact of the detected NRC. We should be able to also classify the root cause of the NRC.

This study considers a set of unique features for each type of NRC and extracts such features from the data to infer the NRC. Thereafter, machine learning models are used to identify the specific type of NRC. Specifically, incidents and workzones are essentially characterized by problematic spots. For inclement weather, we assess the trajectory travel time, speed and gap. And special events are characterised by their impact region and demand surge. The contributions of this paper are:

- Evaluation of machine learning methods for the classification of congestion into its components taking traffic features into account for the inference.
- An algorithm able to detect, identify and propagate via VANET the cause of NRC.
- Validation of the inference methods is made relying on simulation scenarios extended from the real-world Cologne scenario [4].

This paper is organized as follows. Related work is provided in Section II. In Section III, we present our framework. In Section IV, we describe the simulation and provide results. Finally, conclusions and future work are outlined in Section V.

## II. Related Work

Lots of works in congestion detection via VANET utilize machine learning to classify the traffic state into congested or free-flow [5]. To classify the level of congestion, [6] proposes a traffic congestion quantification process based on fuzzy theory. The level has values ranging from free flow to severely congested. These approaches only detect congestion and do not clarify if it is due to recurrent or NRC. The monitoring done by the schemes does not allow summarizing valuable knowledge in an efficient way.

Context-awareness is the potential to access available semantic information such as time, location, weather, temporary events and other attributes [7]. The context information used in [8] fuses different data-sources (internal sensors, web services or passenger sensors) for congestion detection. Their scheme requires additional infrastructure and communication. Without the use of external data sources for inference, our local and self-organized method classifies based on the real-time relevant information extraction by taking advantage of the streaming differentiating characteristic of VANET. Vehicles need to be context aware and able to consider multiple but adequate explanatory sources, well-tailored information won't always be available, particularly in dynamic urban networks. Due to real-time constraints much more information extraction techniques are needed to extract transport-relevant parameters. Statistical inference and machine learning algorithms can provide crucial help in this process. Understanding the causes of urban congestion is a prerequisite for deriving policies and management plans so that appropriate proactive strategies can be set in place.

## III. General Process

The observed travel time of a vehicle ($oTT$) along a road segment may be composed of recurrent delay ($D_{rec}$) and non-recurrent delay ($D_{n-rec}$) such as incident ($D_i$), workzone ($D_{wo}$), weather ($D_{we}$) or special event ($D_{se}$) induced delays.

$$oTT = D_{rec} + (D_i \lor D_{wo} \lor D_{we} \lor D_{se}) \qquad (1)$$

$D_{rec}$ is the expected recurring historic travel time $TT_h$ that is location and time specific. The observed travel time along a segment can be easily obtained by the vehicles of the VANET. If it is higher than a threshold, which is determined as in [9] by multiplying the congestion factor $c$ with the expected recurring delay, the travel time is said to be excessive.

$$oTT > (1+c) * TT_h \Rightarrow oTT \text{ is excessive.} \qquad (2)$$

We claim that real-time traffic flow data collected along a single vehicle trajectory, experience on other road segment and aggregated values offer statistically understandable spatial and temporal features that can help infer the component causing the excessive delay. The resulting classification problem takes the real-time estimates as input feature vector for inference on the cause of congestion. Thus, the general process of our framework is divided into three phases: Features extraction, classification models and cooperative process.

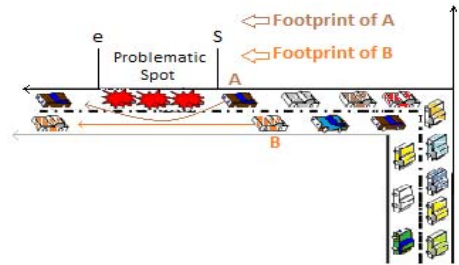

Fig. 1. Trajectory of vehicles around a Pspot.

### Phase 1: Features Extraction

A vehicle can recognize via its neighbours if it's in a jam via cooperative VANET congestion detection. The communication characteristics of a VANET are mostly based on a message called BEACON, transmitted by each vehicle every 0.1 seconds. The message contains time-stamped basic vehicle state information, such as senderID, position, direction, current speed, with optional information also possible. We present below relevant features that characterize each NRC component.

Incidents and workzones are essentially characterized by problematic spots, Pspot. As in [10], we use position data to extract the distribution of vehicle footprints (i.e., the geographical position at each sampling time point on the road). Vehicles periodically register coordinates of their neighbors. If a section of the road is blocked, no position coordinates are recorded between the start and end position of the Pspot as shown in Fig. 1. This feature also considers the temporal aspect of the observed problematic spot. It is a good indicator of an NRC caused by a workzone if the event lasts more than one hour. Often times, workzones occupy the road segment a longer period of time than incidents because incidents are undesired and should be cleared as fast as possible.

NRC caused by inclement weather impacts the trajectory travel time (TrajectoryTravelTime), speed (TrajectorySpeed) and gap (TrajectoryGap). TrajectoryTravelTime measures travel time on the edges upstream in the trajectory of the vehicle and compares them to their respective expected travel time. TrajectorySpeed aims at summarizing speed data along the vehicles' path. TrajectoryGap collects minimum following distances because in inclement weather drivers try to maintain a higher minimum following distance in order to cope with longer stopping distances caused by slippery roads.

Special events are characterized by their impact region (ImpactRegion) and demand surge (TrajectoryDemand). Each segment of the road network is labeled as inside or outside of an impact region. Such an *impact region* can be defined as the list of congested segments of the road network around the special event [11]. We assume that if a vehicle experiences a NRC caused by a special event, then the vehicle is necessarily in the impact region of the event. TrajectoryDemand measures the flow on the edges of a route. Using the speed-flow relationship and knowing the average observed speed on the segment and current density, we estimate flow and compare it with maximum flow, approximated by one fourth of the product of free flow and jam density. When the vehicle detects excessive congestion on a segment, the algorithm ignores the information broadcasted from the trajectory in the last congested segments leading to the excessive congestion segment. The algorithm then only takes into account the flow on the prior segments totaling on average 1.5km. On those upstream segments, traffic is free-flowing and the flow equals the demand. We then compute the weighted average as indictor of demand along a trajectory. Since the demand surge depends on the existing condition of the particular road that is being impacted by the change in demand, we use this feature as an overall sense and feel of the road segment in comparison to the maximum flow denominator that makes this measure comparable. The procedure starts with every vehicle measuring the flow on each road segment of its trajectory.

Finally, CurrentTT is a feature that categorizes the travel time observed along a segment as normal or excessive according to Equation (2). Also, excessive travel time can be noted on a road segment adjacent to one where the congestion was initially detected. Cooperation between vehicles can propagate the event to adjacent roads. A StoredEvent feature might indicate that incident, workzone, special event, weather, or that no stored event on the segment exists. The vector of features is provided as input to the classification models for inference on the cause of the NRC.

### Phase 2: Classification Models

Tree models where the target variable can take a finite set of values are called classification trees [12]. C4.5 is an algorithm used to build classification trees from a set of training data using the concept of information entropy [12]. The purpose is to split at each node with the feature having the highest normalized information gain. We employ such an algorithm in the training of the CT described in this paper.

We also developed a probabilistic model based on a Naive Bayesian classifier which gives useful predictions about the congestion. The aim of the Naive Bayesian classifier is to assign a target variable to one of a discrete set of categories based on its observable features.

$$P(Y|x) = \frac{P(Y)P(x|Y)}{P(x)}$$

Applied to our problem, translation of the a posteriori observable characteristics $x_1, x_2, \ldots, x_i$ into congestion component class $I$ of $y_1, \ldots, y_j$ is computed by using Bayes rules:

$$P(I \in Y_j | x_1, \ldots, x_i) = \frac{P(I \in Y_j)P(x_1, \ldots, x_i | I \in Y_j)}{P(x_1, \ldots, x_i)}$$

The classifier is naive because it makes the strong assumption that the features are mutually conditionally independent; that is, the conditional probability that $I$ belongs to a particular class given the value of some feature is independent of the values of all other features. There is no statistically significant data for assessment of the more complicated causality between explanatory variables. Also, since the parameters of the NB model are estimated, probabilistic dependencies among features need contextual observations, the lack of ground-truth data prevents this research from fully modeling the realism of this transport-related phenomena. Despite this assumption, empirical studies demonstrate that it does not significantly compromise the accuracy of the prediction. This reduces the probability to:

$$P(I \in Y_j | x_1, \ldots, x_i) = \frac{P(I \in Y_j) \prod\limits_{z=1}^{i} P(x_z | I \in Y_j)}{P(x_1, x_2 \ldots, x_i)}$$

$I$ is typically assigned to the category with the greatest probability. The most likely $j$ is chosen as follows:

$$j^* \in \arg\max P(I \in Y_j) \prod_{z=1}^{i} P(x_z | I \in Y_j) \qquad (3)$$

and assigning $I$ to class $Y_{j*}$.

### Phase 3: Cooperative Process

If excessive travel time is detected on a segment, the scheme activates a cooperative process that shares the individual estimation made by the vehicle. We present in Fig. 2 the algorithm implemented on board of each vehicle and that is primarily for signalized arterials. We highlight the monitoring, aggregating, analysis and dissemination procedures. The purpose is to assess if the temporary induced traffic change related to an event can be mitigated in a short period or does the event represent a permanent change representing an NRC. We implemented the methods described in [2], Basic Traffic Data Gathering algorithm, Local Traffic Evaluation Algorithm and Expanding the Evaluated Area algorithm. We add two types of messages: Extraordinary Event Request (RQ) and Extraordinary Event Response (RP). RQ is transmitted upstream via broadcast to all cars in its communication range, and allow to retain the event info locally on the segment for a minimum duration before propagating it to adjacent segments. In [9], it was shown that continuously high values of travel time along a segment that last at least four successive time intervals was the criterion used as evidence of a NRC event. This prevents false positive NRC detection. RP is the message used to send the NRC event to adjacent segments after the duration expires.

## IV. SIMULATION

The method provided in our study is applied to a heterogeneous network of both urban highways and signalized arterials [9]. The real-world traffic dataset of the Travel and Activity PAtterns Simulation (TAPAS) Cologne scenario [4] is considered a 'complex network' that mimics the real-life context of vehicle mobility. Heterogeneity exists on urban road networks where the structure of links varies substantially. The dataset comprise 700 000 individual car trips. Each line of the dataset contains the time, the vehicle identifier, its position and speed. Using SUMO, a microscopic traffic simulator for the simulation of urban mobility [13], we create extended scenarios mounted on top of the base scenario to model atypical traffic conditions such as weather, incident, workzone and special event. SUMO simulator needs two inputs: The Road Network of the city of Cologne is imported from the OpenStreetMap (OSM) database and the Traffic Demand is the dataset of car trips. The output of SUMO is the movement

**DATA** - *Vi*: Vehicle in the scenario, *oTT*: Observed travel time, *TTh*: Historical travel time, *CurrentTT(Vi)*: Travel time and local traffic evaluation, *TrajectoryTT(Vi)*: Travel time on edges of route stored in EdgesofRouteofV, *TrajectorySpeed(Vi)*: Speed of vehicles stored in ListeEdges, *TrajectoryDemand(Vi)*: Flows in ListeEdgesD, *TrajectoryGap(Vi)*: Gap distances in vehiclesG.

1: *Vi* broadcasts and recieves BEACON message from neighbors // MONITORING
2: Get current road segment of *Vi* and CurrentTT(*Vi*) on the segment // AGGREGATING
3: Update TrajectoryTT(*Vi*), TrajectorySpeed(*Vi*), TrajectoryDemand(*Vi*) and TrajectoryGap(*Vi*)
4: **if** *oTT* > 1.8 * *TTh* **then** // ANALYSIS and DISSEMINATION
5:     Calculate features
6:     Create feature vector and Predict with BN
7:     **if** StoredEvent == 0 **then**
8:         *Vi* creates and backpropagates RQ
9:     **else**
10:         **if** Duration not reached **then**
11:             Store RQ
12:         **else**
13:             **if** Duration reached or NRC is Incident or Workzone **then**
14:                 Backpropagate RP to adjacent road segments
15:             **end if**
16:         **end if**
17:     **end if**
18: **end if**

Fig. 2.   Algorithm–cooperative process of VANET.

of vehicular nodes in a large urban network and data such as the acceleration, density, flows, gap between vehicles and other microscopic parameters at a vehicle level. From the simulation data collected by each vehicle, we extract features constituting an instance of the train dataset.

To simulate the Extended Scenarios of an Incident/ Workzone, on the base scenario, we stop on a lane some vehicles for a specific amount of time. We vary the position on the edge and the duration. For the Extended scenario of bad weather, which lead to decreases in the vehicles' velocities and a more careful and defensive driver behaviour, we change the parameters of the car-following model in the simulator. For the special event scenario, to generate trips to a particular destination, we generate random departures and random routes. We use a Poisson process to generate random timings for trips. The rate parameter $\lambda$ is the demand per second from different sources. To generate random routes, given trips are assigned to respective fastest routes according to their departure times and a given travel-time updating interval by SUMO's traffic assignment model.

We construct a training dataset, a matrix with rows corresponding to samples and columns to features. The train dataset contains 591 instances. To obtain a realistic environment for the simulation of vehicular communications, we extract from the extended scenarios in SUMO, the vehicular traces that we will use in ns2 [14]. We assume that vehicles are equipped with a Global Positioning System (GPS) device for positioning, a transceiver for communication using Dedicated Short-Range

Communications (DSRC), and an enriched digital road map containing information about the map, including the length of each road, number of lanes per road, SegmentID and the expected travel times on the segments. We use data forwarding techniques to pass information through the VANET such as geographical routing, and broadcast. For communication among all cars, we assume standard signal range of the 802.11p protocol, which is 300 meters.

### A. Results

We demonstrate the robustness of our scheme by examining the performances of accuracy of classification, timing, and impact of NRC in an urban network. Firstly, we use Weka to generate a pruned classification tree [15]. Weka is a suite of machine learning software for data analysis and predictive modeling. The proposed CT is presented in Fig. 3. The accuracy of classification measures the predictive performance of the classifier and is determined by the percentage of the test dataset examples that are correctly classified. We performed 10 fold cross-validation on the training set and we got 87.63% of correctly classified instances. The value ranges of the splitting arcs are learned by the classifier and shown as nominal values on the arcs of the tree. The tree starts with Current-TravelTime feature as the root node. CurrentTravelTime on a segment measures the travel time of a vehicle on each segment and compares it with $TT_h$ of each segment. CT confirmed that when travel time on a segment is below its excessive treshold, the congestion is due to recurrent congestion. Otherwise, it is a NRC and if there is a problematic spot on the segment, the tree attributes the NRC cause to either an incident or a workzone. Then, the tree splits on the StoredEvent feature. If the vehicle is in a congestion due to a special event, its location is necessarily inside the impact region of the event. The tree splits on the internal non-leaf node labeled ImpactRegion. The leaf node SpecialEvent is on branches coming out of inside an impact region. TrajectoryDemand and TrajectoryGap diffrentiate data between a special event and a weather condition. Problems of small-data mainly revolve around high variance were outliers are present. More training cases are needed for the statistical inference to pick up the causality between explanatory features in order to make strong assumption.

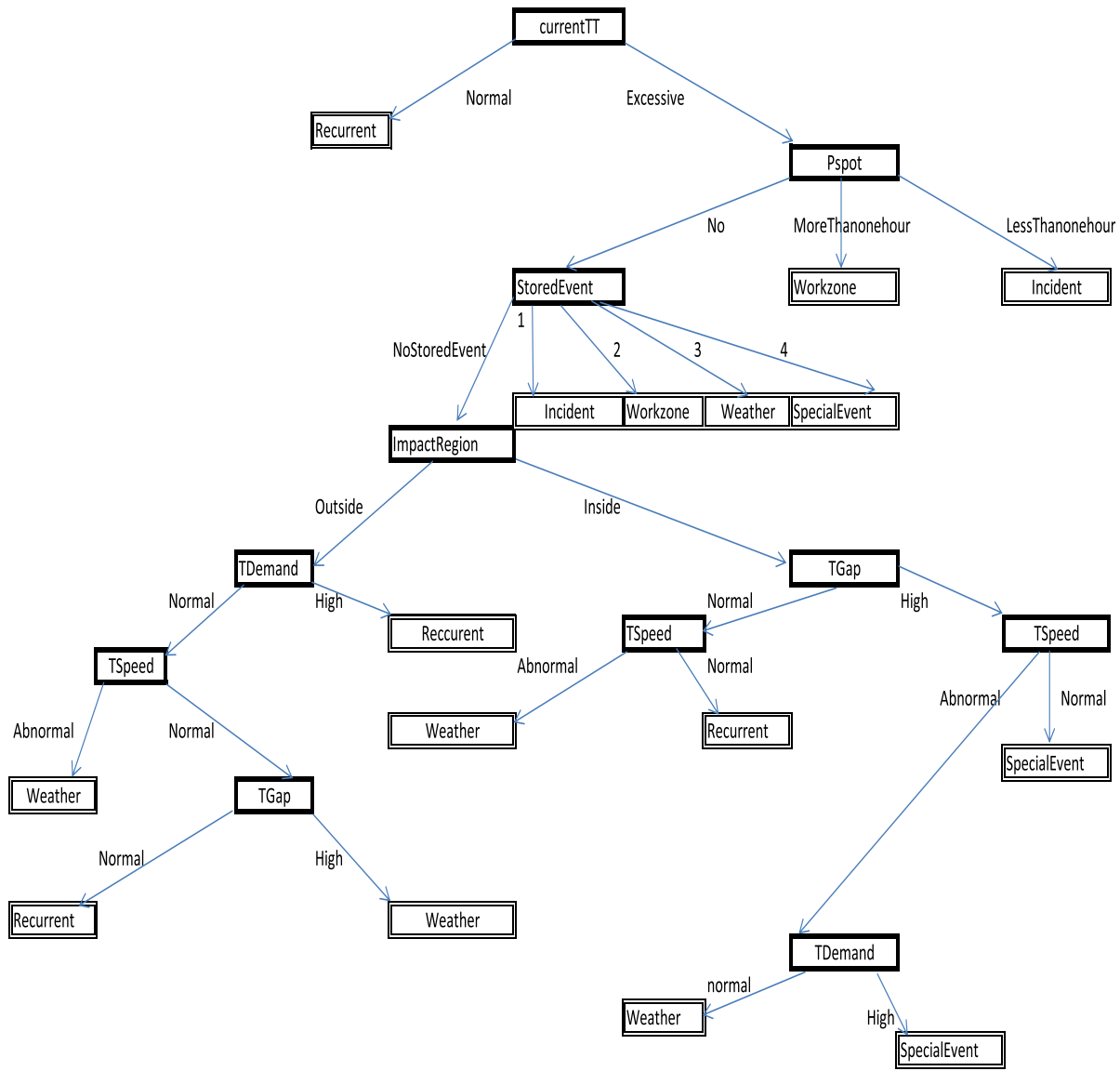

Fig. 3. Classification tree.

BN is presented in Fig. 4. We performed 10 fold cross-validation on the training set to test the model. It's accuracy in terms of prediction error is 88.83% of correctly classified instances. Dependencies between a cause of congestion and its consequences are represented by arcs on the graph. All causes of congestion, except the recurrent congestion, have arcs going to ImpactRegion and SoredEvent features. This is because any NRC might occur inside an impact region as well as outside. Also, a StoredEvent feature will rule out all other causes of congestion if its value reports workzone, incident, weather or special event. We conducted a sensitivity analysis on the features of the model to note the importance of a feature for some partial classification. We removed one feature at a time and used the filtered training set for classification. We show in Fig. 5 the sensitivity of each feature on the accuracy of the CT and BN. We see that except for StoredEvent, the other features have the same performance.

A possible extension of the CT method described in our paper is Random forests. It's an ensemble learning method also used for classification. It's combining multiple models into ensembles to produce an ensemble for learning. It operates by constructing a multitude of decision trees at training time and outputting the class that is the mode of the classes of the individual trees. As in CT, we use J48 to produce decision trees, then we produce slightly different decision trees by randomization. Weka's implementation yielded 89.51% accuracy of Random forest of 100 trees, each constructed while considering 4 random features.

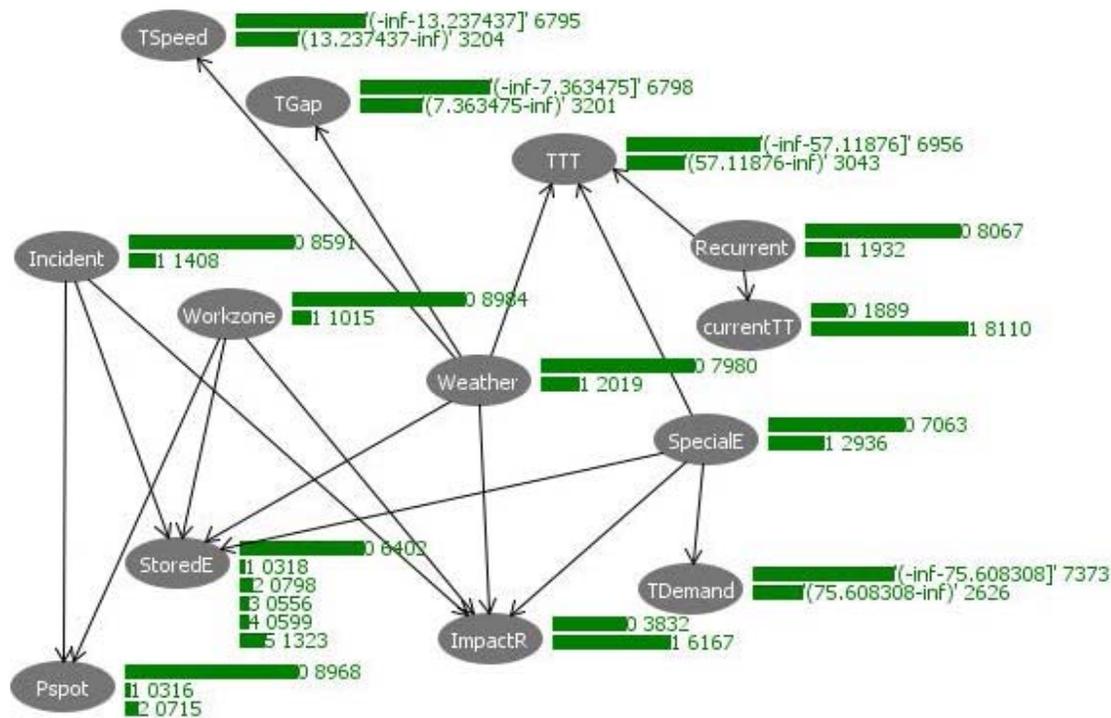

Fig. 4.   Bayes network.

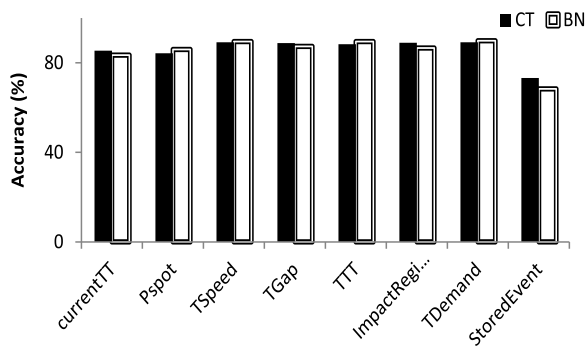

Fig. 5.   Sensitivity of the CT and NB models.

Boosting is a fairly recent technique in supervised learning. AdaBoostM1 is a standard boosting scheme where diversity is created by focusing on where the existing model makes errors. Iteratively, new models are influenced by the performance of previously built models. Extra weight is given to instances that are misclassified to make a training set for producing the next model in the iteration. This encourages the new model to become an 'expert' for instances that were misclassified by all the earlier models. AdaBoostM1 is implemented in Weka. With 10 iterations, it yielded 89.17% of accuracy. In an urban road network context, the models were able to classify the cause of the congestion on both highways and signalized arterials because classification is based on the collection of features (problematic spot, currentTT, etc.) that are nonspecific to the type of the facility.

Secondly, vehicles in our scheme are able to monitor traffic and detect NRC during different times. In Fig. 6, we illustrate average travel time (TT1-TT4) of vehicles during incidents happening at different times (T1-T4 correspondingly) on the same busy road segment. Any time congestion is detected and the average travel time is above ExcessiveTT, we assess the number of vehicles reporting severe congestion compared to the number of vehicles on the segment. We highlight on Fig. 6 the time when congestion from incidents happens and denote it by TC. TC1 is related to the incident happening at T1 because the time of congestion differs from the time of the incident. The level of excessive travel time induced by the incidents can be observed in the figure but it's only when congestion is detected at TC and values are higher than the threshold determined with the congestion factor that NRC is declared. Also, as in [9], continuously high values for at least four successive time intervals was the other criterion used as evidence of an event. The results indicate that on average, 88% of vehicles were able to detect the NRC and the percentage gradually increased to 95% in the next 5-15 minutes interval. We conclude that severe delay caused by NRC can be accurately detected any time it happens. To guarantee monitoring of road traffic at various level of granularity, a variable congestion factor could be considered interactively for management purposes. But for NRC detection, studies showed that a fixed congestion factor can accurately detect most NRC [9], as was the case in our experiments.

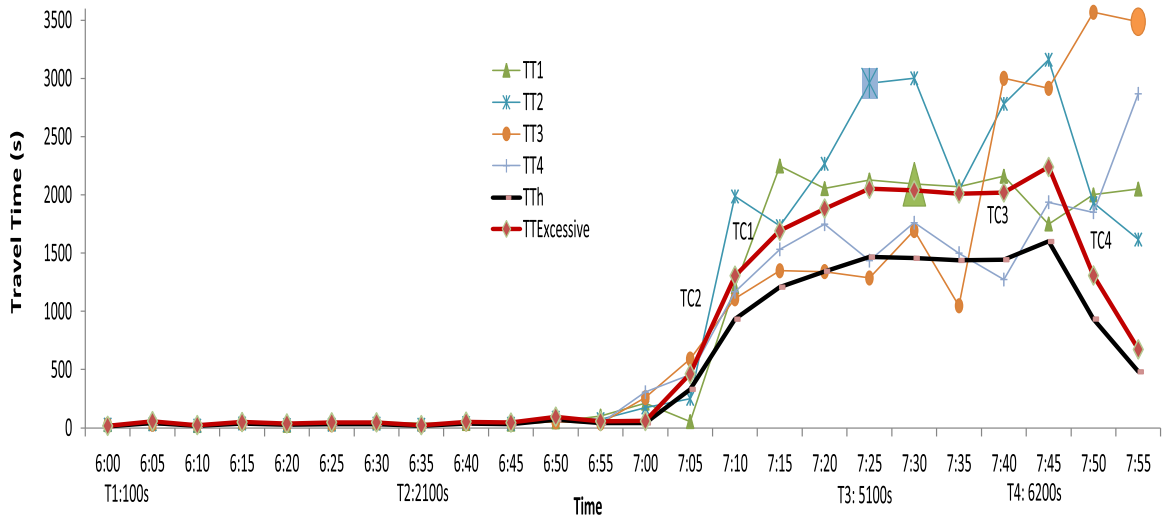

Fig. 6.  Average Travel Time on a signalized arterial during incidents happening at times T1 to T4.

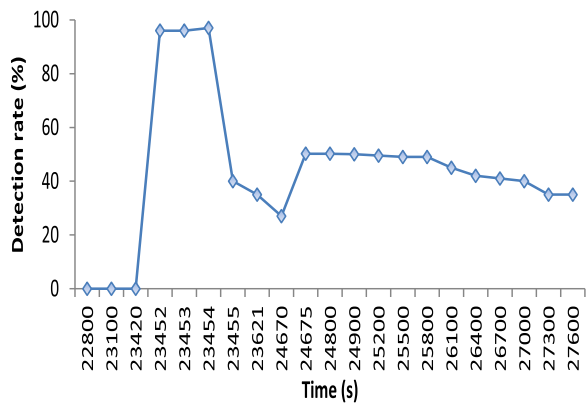

Fig. 7.  Accuracy of the impact region.

Finally, we demonstrate that vehicles are able via VANET to propagate the cause of NRC. For special events, we tested the accuracy of the impact region. We report the percentage of vehicles experiencing NRC due to a special event inside the impact region in contrast to all those experiencing the congestion from the special event, inside and outside the predefined impact region of the event. We monitor an impact region of only one road segment and we note that the size of the impact region has an initial impact on the detection rate, as seen in Fig. 7.

Congestion from the special event starts around t = 22100s but it's only at t = 23422s, that it becomes excessive. Vehicles inside the impact region are the first to detect the congestion and its cause and the detection rate is very high. After, the detection gradually decreases to 27.58% and increases again to 50.24% at t = 24670s. This behavior is due to the cooperative process of our method. Communication between vehicles on the same segment has to happen for a certain duration before propagation of the event to first order adjacent segments can be done. During this period, more vehicles outside the impact region are experiencing severe congestion from the special event but cannot accurately assess the cause, for this reason, detection rate decreased. Then, detection rate increased when more vehicles outside the region became aware of the event after communication between segments via VANET. The algorithm stops after the first-order adjacent segments; consequently, detection rate cannot get any higher. The scheme does not evaluate the spatial extent of the NRC. It is out of the scope of this study. But the demonstration showed that the ImpactRegion feature can be used without any knowledge about the event because the region is reshaped via communication.

## V. CONCLUSION

The duration, timing and location of non-recurrent congestion (NRC) in an urban network varies a lot making it difficult to monitor traffic in real time with conventional mechanisms. We have proposed a framework for the distributed classification of congestion into its components using VANET as an alternative cost-effective and flexible solution to guarantee better monitoring of road traffic on heterogeneous networks. The proposed framework aims to exchange traffic flow data and to embed reasoning machinery in vehicles to infer the cause of NRC.

We have obtained a predictive accuracy of 87.63% for the classification tree (CT), 88.83% for the Bayesian network (BN), 89.51% for Random forest (RF) and 89.17% for the boosting technique trained on synthetic data extended from the real case study of the Cologne scenario. In the future, we note that more sophisticated methods can be employed in the cooperative process, such as a voting process, a likelihood evaluation or a model of the value of information. Also, data of connected vehicle operations in real-world conditions, such as Ann Arbor Automated Vehicle Operational Test, can be used as a test environment and provide real-world training dataset in occurance of different NRC scenarios [16].

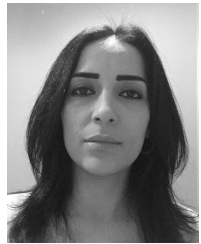

**Ranwa Al Mallah** received the M.Sc. degrees in computer science from the École Polytechnique de Montréal, Canada. She is currently working toward the Ph.D. degree with the LARIM Research Laboratory, University of Montreal. Her research interest includes vehicular networks, traffic efficiency, and safety applications.

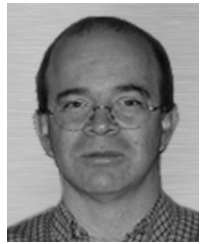

**Alejandro Quintero** received the B.E. degree in computer engineering from University of Los Andes, Colombia, in 1983; the Diploma degree in advanced studies from Grenoble INP, Grenoble, France, in 1993; and the Ph.D. degree in computer engineering from Joseph Fourier University, Grenoble. He is a Full Professor with the Department of Computer Engineering, École Polytechnique de Montréal, Canada. He has co-authored two books and over 100 other technical publications, including journal and proceedings papers.

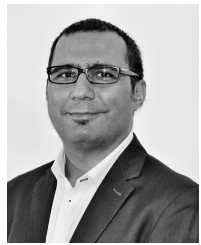

**Bilal Farooq** was born in the foothills of the Himalayas in Pakistan. He received the B.E. degree from University of Engineering and Technology, Pakistan, in 2001; the Master in Computer Science degree from Lahore University of Management Sciences, Pakistan, in 2004; and the Ph.D. degree from University of Toronto, Canada, in 2011. His research goal is to develop multidisciplinary and highly-intelligent solutions for sustainable planning, design, and operations of urban infrastructure systems.